\documentclass{article} 
\usepackage{nips13submit_e,times}
\usepackage{hyperref}
\usepackage{url}
\usepackage{amsfonts}
\usepackage{graphicx}
\usepackage{subfigure}

\usepackage[boxed,noend]{algorithm2e}
\title{Generative NeuroEvolution for Deep Learning}

\author{
Phillip Verbancsics \& Josh Harguess \\
Space and Naval Warfare Systems Center -- Pacific\\
San Diego, CA 92101\\
\texttt{\{phillip.verbancsics,joshua.harguess\}@navy.mil} \\\
}

\nipsfinalcopy 

\begin{document}

\maketitle

\begin{abstract}

An important goal for the machine learning (ML) community is to create approaches that can learn solutions with human-level capability. One domain where humans have held a significant advantage is visual processing. A significant approach to addressing this gap has been machine learning approaches that are inspired from the natural systems, such as artificial neural networks (ANNs), evolutionary computation (EC), and generative and developmental systems (GDS). Research into deep learning has demonstrated that such architectures can achieve performance competitive with humans on some visual tasks; however, these systems have been primarily trained through supervised and unsupervised learning algorithms. Alternatively, research is showing that evolution may have a significant role in the development of visual systems. Thus this paper investigates the role neuro-evolution (NE) can take in deep learning. In particular, the Hypercube-based NeuroEvolution of Augmenting Topologies is a NE approach that can effectively learn large neural structures by training an indirect encoding that compresses the ANN weight pattern as a function of geometry. The results show that HyperNEAT struggles with performing image classification by itself, but can be effective in training a feature extractor that other ML approaches can learn from. Thus NeuroEvolution combined with other ML methods provides an intriguing area of research that can replicate the processes in nature.

\end{abstract}

\section[Intro]{Introduction}


\indent Evolution has a significant role in creating biological visual systems \cite{le::2013,gliga::2005}, thus a potential path for training artificial vision as effective as the biological counterparts is neuro-evolution (NE; \cite{yao:ieee99}) . However, NE has been challenged because of the required structure sizes. While the human brain includes trillions of connections \cite{kandel:book00,neuroscience99}, traditional NE approaches produce networks significantly smaller in size \cite{yao:ieee99,stanley:alife09}. Thus an active area of research is evolutionary approaches that address this gap \cite{stanley:alife09}. One significant area of interest is generative and developmental systems (GDS), which are inspired by principles that map genotype to phenotype in nature. Such principles are enabled by reusing genetic information \cite{kandel:book00,bentley:gecco99}. For example, biological neural networks exhibit several important organizational principles, such as modularity, regularity, and hierarchy \cite{hartwell:nature99,wagner:evolution96}, which can be exploited through the indirect encoding and enable evolution of complex structures.

\indent The GDS approach investigated in this paper is called the Hypercube-based NeuroEvolution of Augmenting Topologies (HyperNEAT; \cite{stanley:alife09,gauci:aaai08,gauci:nc10}). HyperNEAT trains an indirect encoding, called a \emph{compositional pattern producing network} (CPPN; \cite{stanley:gpem07}) that represents artificial neural network (ANN) connection weights. HyperNEAT has shown promise in simple visual discrimination tasks \cite{gauci2007,coleman::honors2010,verbancsics:gecco11,Hausknecht2012}, but has been limited to two or fewer hidden layers. This paper extends the investigation to deeper architectures by applying HyperNEAT to the MNIST benchmark dataset. Interestingly, the results show that HyperNEAT alone has difficulty in classifying the images, but HyperNEAT is effective at training a feature extractor for backward propagation learning. That is, HyperNEAT can learn the `deep' parts of the neural network, while backward propagation can perform the fine tuning to make classifications from the generated features. 

\section[Back]{Background}

\indent The geometry-based methods that underlie the approach are described in this section.

\subsection{NeuroEvolution of Augmenting Topologies (NEAT)}

\indent \emph{Neuro-evolution} (NE; \cite{yao:ieee99}) methods train artificial neural networks (ANNs) through evolutionary algorithms. As opposed to updating weights according to a learning rule, candidate ANNs are evaluated in a task and assigned fitness to allow selection and creation of a new generation of ANNs by mutating and recombining their selected genomes. The NeuroEvolution of Augmenting Topologies (NEAT) algorithm \cite{stanley:jair04} is a popular neuroevolutionary approach that has been proven in a variety of challenging tasks, including particle physics \cite{aaltonen:prl09,whiteson:iaai07}, simulated car racing \cite{card::cec2009}, RoboCup Keepaway \cite{taylor:gecco06}, function approximation \cite{whiteson:aamas05}, and real-time agent evolution \cite{stanley:ieeetec05}, among others \cite{stanley:jair04}.

\indent NEAT starts with a population of small, simple ANNs that increase their complexity over generations by adding new nodes and connections through mutation. That way, the topology of the network does not need to be known a priori; NEAT searches through increasingly complex networks as it evolves their connection weights to find a suitable level of complexity. The techniques that facilitate evolving a population of diverse and increasingly complex networks are described in detail in Stanley and Miikkulainen \cite{stanley:jair04}; the important concept for the approach in this paper is that NEAT is an evolutionary method that discovers the right topology and weights of a network to maximize performance on a task. The next section reviews the extension of NEAT called HyperNEAT that allows it to effectively train large neural structures.

\subsection{CPPNs and HyperNEAT}

\indent Hypercube-based NEAT (HyperNEAT; \cite{stanley:alife09,gauci:nc10}) is a GDS extension of NEAT that enables effective evolution of high-dimensional ANNs. The effectiveness of the geometry-based learning in HyperNEAT has been demonstrated in multiple domains, such as multi-agent predator prey \cite{dambrosio:aamas2010,dambrosio:gecco08} and RoboCup Keepaway \cite{verbancsics:jmlr10}. A full description of HyperNEAT is in Stanley et al.\ \cite{stanley:alife09}.

\indent The main idea in HyperNEAT is that geometric relationships are learned though an indirect encoding that describes how the \emph{weights} of the ANN can be \emph{generated} as a function of geometry. Unlike a direct representation, wherein every connection in the ANN is described individually, an indirect representation describes a pattern of parameters without explicitly enumerating each such parameter. That is, information is reused in such an encoding, which is a major focus in the field of GDS from which HyperNEAT originates \cite{embryogeny03,turing52}. Such information reuse allows indirect encoding to search a compressed space. HyperNEAT discovers the \emph{regularities} in the geometry and learns from them.

\begin{figure}[t]
\begin{center}
\centering
\includegraphics[width=0.65\linewidth]{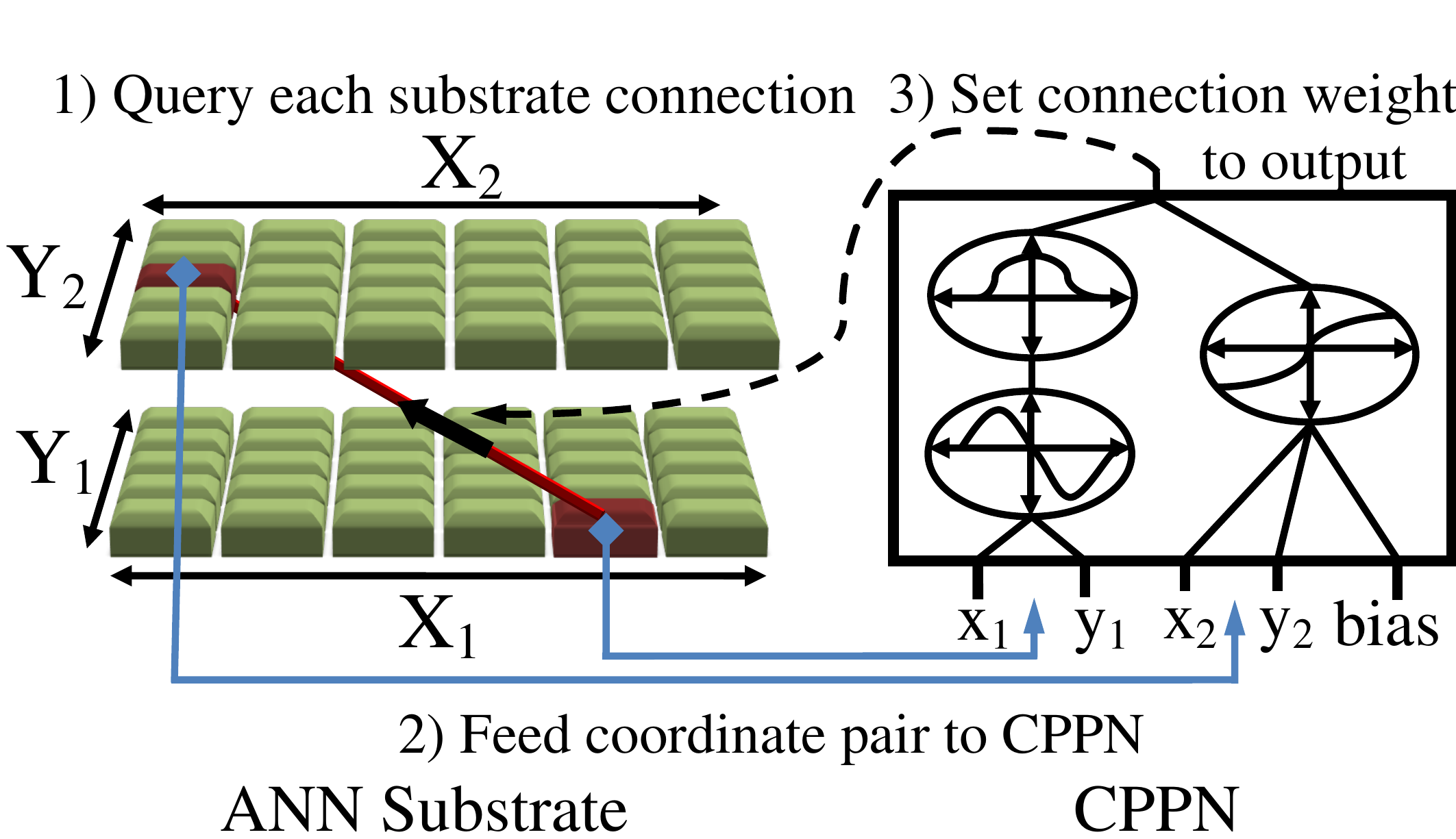}
\end{center}
\vspace{-12pt}
\caption{\textbf{A CPPN Describes Connectivity.}
A grid of nodes, called the ANN \emph{substrate}, is assigned coordinates. (1) Every connection between layers in the substrate is queried by the CPPN to determine its weight; the line connecting layers in the substrate represents a sample such connection. (2) For each such query, the CPPN inputs the coordinates of the two endpoints. (3) The weight between them is output by the CPPN. Thus, CPPNs can generate regular patterns of connections.
\vspace{0.0in}
}
\label{cppn:}
\end{figure}

\indent The indirect encoding in HyperNEAT is called a \emph{compositional pattern producing network} (CPPN; \cite{stanley:gpem07}), which encodes the \emph{weight pattern} of an ANN \cite{stanley:alife09,gauci:aaai08}. The idea behind CPPNs is that geometric patterns can be encoded by a \emph{composition of functions} that are chosen to represent common regularities. 
In this way, a set of simple functions can be composed into more elaborate functions through hierarchical composition.
Formally, CPPNs are \emph{functions} of geometry (i.e.\ locations in space) that output connectivity patterns for nodes situated in $n$ dimensions. 
Consider a CPPN that takes four inputs labeled $x_1$, $y_1$, $x_2$, and $y_2$; this point in four-dimensional space can \emph{also} denote the connection between the two-dimensional points $(x_1, y_1)$ and $(x_2, y_2)$. The output of the CPPN for that input thereby represents the weight of that connection (figure \ref{cppn:}).

\indent Because the connection weights are produced as a function of their endpoints, the final pattern is produced with \emph{knowledge} of the domain geometry, which is literally depicted geometrically within the constellation of nodes. Weight patterns produced by a CPPN in this way are called \emph{substrates} so that they can be verbally distinguished from the CPPN itself. It is important to note that the structure of the substrate is independent of the structure of the CPPN. The substrate is an ANN whose nodes are situated in a coordinate system, while the CPPN defines the connectivity among the nodes of the ANN. The experimenter defines both the location and role (i.e.\ hidden, input, or output) of each node in the substrate.


\indent In summary, HyperNEAT evolves the topology and weights of the CPPN that \emph{encodes} ANN weight patterns. An extension of HyperNEAT called HyperNEAT with Link Expression Output (HyperNEAT-LEO) was introduced to constrain connectivity with a bias towards modularity \cite{verbancsics:gecco11}. This extension separates the decision of weight magnitude and expression into \emph{two} different CPPN outputs and seeds the LEO with the concept of locality. The HyperNEAT-LEO variant is shown in Algorithm 1. The next section reviews deep learning that has been applied to image classification.

\begin{algorithm}
\KwIn{Substrate Configuration}
\KwOut{Solution CPPN}
Initialize population of minimal CPPNs with random weights\;
\While{Stopping criteria is not met}{
\ForEach{CPPN in the population}{
\ForEach{Possible connection in the substrate}{
Query the CPPN for weight $w$ of connection and LEO expression
\vspace{0pt} value $e$\;
\If{$e > 0.0$}{
Create connection with a weight $w$\;
}
}
Run the substrate as an ANN in the task domain to ascertain
\vspace{0pt} fitness\;
}
Reproduce CPPNs according to the NEAT method to produce the
\vspace{0pt} next generation\;
}
Output the champion CPPN.
\label{HyperNEATalg}
\caption{HyperNEAT-LEO Algorithm}
\end{algorithm}

\subsection{Deep Learning in Image Classification}

\indent Neural networks have experienced a resurgence thanks to breakthroughs in deep learning that have led to state of the art results in a number of challenging domains \cite{Bengio2013}. In particular, deep learning approaches have achieved remarkable performance in a number of object recognition benchmarks, often achieving the current best performance on these tasks. Such object recognition tasks where deep learning has achieved the best results include the MNIST hand-written digit dataset \cite{Hinton2006,Rifai2011}, traffic sign recognition \cite{Ciresan2012}, and the ImageNet Large-Scale Visual Recognition Challenge \cite{Krizhevsky2012}.

\indent The challenge for deep learning is how to effectively train such large neural structures. Traditional supervised learning approaches for neural networks, such as backward propagation, face problems that include the ``curse of dimensionality'' or vanishing gradient. Thus a popular alternative is to perform pre-training on the deep architecture through unsupervised learning. Restricted Boltzmann Machines \cite{Salakhutdinov2009} and auto-encoders \cite{Hinton2006a} are among the common approaches to performing this unsupervised training. This pre-training can either act as an initial starting point for supervised learning of a deep network or a feature extractor from which machine learning approaches can learn \cite{Bengio2013}.

\indent Alternatively, supervised learning performance can be enhanced through the selection of the deep architecture, such as \emph{convolutional neural networks} (CNNs). Beginning with the Neocognitron \cite{Fukushima1982}, these deep architectures have been inspired by proposed models of the human visual cortex \cite{Hubel1959} and the concept of local receptive fields that take advantage of the input topology. That is, CNNs enforce a particular geometric knowledge by constructing an architecture that learns features based upon locality. Through local receptive fields, shared weights, and sub-sampling \cite{LeCun1998}, CNNs enable backward propagation (supervised learning) to effectively train deep architectures. The next section introduces approaches to deep learning through HyperNEAT.

\section{Approach: Deep Learning HyperNEAT}

\indent Although the HyperNEAT method succeeds in a number of challenging tasks \cite{stanley:alife09,gauci:aaai08,gauci:nc10,verbancsics:jmlr10} by exploiting geometric regularities, it has not yet been applied to tasks where deep learning is showing promise or to deep architectures. Because HyperNEAT learns as a function of domain geometry, it is well-suited towards architectures similar to CNNs. This approach introduces two modifications to HyperNEAT training.

\begin{figure}[t]
\begin{center}
\centering
\includegraphics[width=0.55\linewidth]{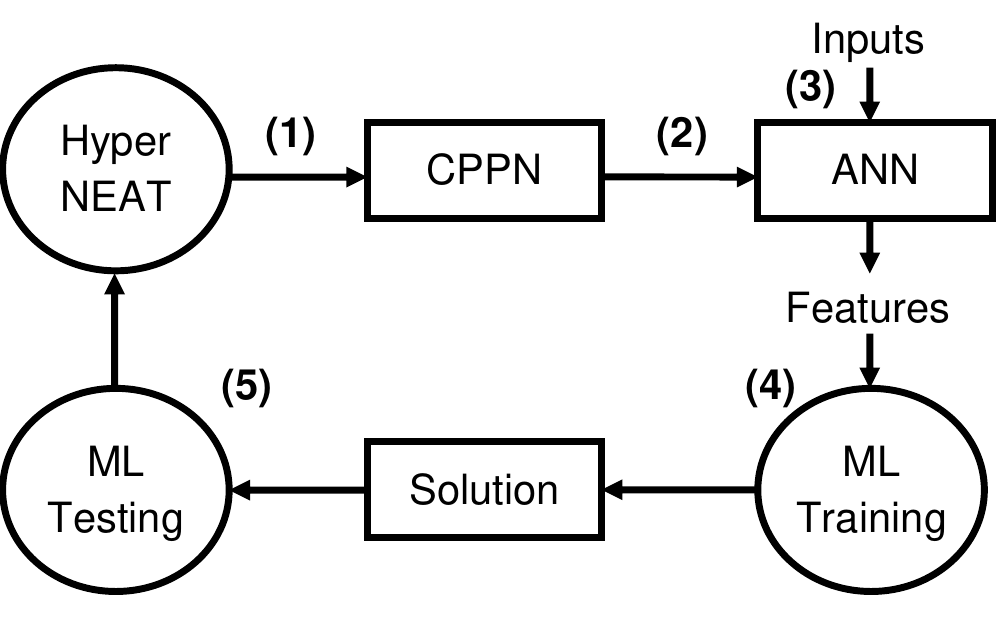}
\end{center}
\vspace{-12pt}
\caption{\textbf{HyperNEAT Feature Learning.} To learn features, HyperNEAT trains CPPNs (1) that generate the connectivity for a defined ANN substrate (2). The ANN substrate processes the inputs from a data set to produced a set of features (3). These features are given to another machine learning algorithm (4) that learns to perform the task (e.g.\ image classification). Machine learning then produces a solution that is evaluated on testing data (5). The performance of the solution on data not seen during training provides the fitness score of the CPPN for HyperNEAT. In this way, HyperNEAT not only discovers better learning features, but also better features for generalization.
\vspace{0.0in}
}
\label{hyperneatfl:}
\end{figure}

\indent The first modification introduces the idea of HyperNEAT as a feature learner. Conventional HyperNEAT trains a CPPN that defines an ANN that is the solution, that is, the produced ANN is applied directly to the task and then the ANN's performance on that task determines the CPPN's fitness score. However, this HyperNEAT modification trains an ANN that transforms inputs into features based upon domain geometry and then the features are given to another machine learning approach to solve the problem, the performance of this learned solution then defines the fitness score of the CPPN for HyperNEAT (figure \ref{hyperneatfl:}). In this way, HyperNEAT acts as a reinforcement learning approach that determines the best features to extract for another machine learning approach to maximize performance on the task. 

\indent The second modification introduces alternative architectures to HyperNEAT. Traditionally, HyperNEAT produces the weight pattern for a ANN substrate that is feed forward, fully connected, and containing only sigmoid activation functions. However, the only information HyperNEAT sees about the substrate is the geometric coordinates of the neurons, thus HyperNEAT could be applied to any graph like structure wherein the nodes have coordinates associated with them. Because CNNs have been demonstrably successful in a number of domains, HyperNEAT is extended to include this alternative ANN architecture. Each of these extensions to HyperNEAT is explored in experiments described in the next section. 

\section{Experimental Setup}

\indent These investigations are conducted on the MNIST dataset, which is a popular benchmark for machine learning because it is a challenging and relevant real world problem. MNIST is a set of $28\times28$ pixel images of handwritten digits ($0-9$), separated into 60,000 training images and 10,000 testing images. The goal for machine learning is to correctly classify the handwritten digit contained within each image. To this end, we explore HyperNEAT with four combinations of experimental settings on the benchmark MNIST dataset. These settings are traditional HyperNEAT ANN architecture (feed-forward, fully-connection, sigmoid activation functions) or CNN architecture and HyperNEAT training the solution (an ANN for classification) or feature extractor (ANN that transforms images into features).

\indent The architecture for HyperNEAT in these experiments is a multi-layer neural network wherein the layers travel along the $z$-axis, each layer consists of a number of features ($f$-axis), and each feature has a constellation of neurons on the $x,y$-plane corresponding to pixel locations. Thus the CPPN represents points in an eight-dimensional Hyper-cube that correspond to connections in the four-dimensional substrate and each neuron is located at a particular ($x,y,f,z$) coordinate. Each layer is presented by a triple ($X,Y,F$), wherein $F$ is the number of features and $X,Y$ are the pixel dimensions. Thus the input layer is ($28,28,1$), because the input image is $28\times28$ and contains only grayscale pixel values. The traditional HyperNEAT architecture for these experiments is a seven layer neural network with one input, one output, and five hidden layers, represented by the triples ($28,28,1$), ($16,16,3$), ($8,8,3$), ($6,6,8$), ($3,3,8$), ($1,1,100$), ($1,1,64$), and ($1,1,10$). Each layer in the traditional HyperNEAT ANN architecture is fully connected to the adjacent layers and each neuron has a bipolar sigmoid activation function. The CNN architecture to which HyperNEAT is applied replicates the LeNet-5 architecture that was previously applied to the MNIST dataset \cite{LeCun1998}.

\indent To operate as a feature extractor, the above architectures are modified such that the ANN substrate is cut off before the last hidden layer, that is, for the traditional ANN architecture the ($1,1,100$) layer becomes the new output layer and for the CNN architecture the ($1,1,120$) layer becomes the outputs. Thus each image is passed through the ANN substrate architecture to produce an associated feature vector. These feature vectors are then given to backward propagation to train an ANN with an architecture identical to the architecture that was removed from the substrates. The next section presents the results of these HyperNEAT variants.

\section{Results}

\indent For each of these experiments, results are averaged over 30 independent runs of 2500 generations with a HyperNEAT population size of 256. The fitness score is the sum of the true positive rate, true negative rate, positive predictive value, negative predictive value, and accuracy for each class plus the fraction correctly classified overall and the inverse of the mean square error from the correct label outputs. Each run randomly selects 300 images, evenly spread across the classes, from the MNIST training set for training. For regular HyperNEAT (i.e.\ not feature learning), fitness is determined by applying the ANN substrate to the training images. For HyperNEAT feature learning an additional 1000 images are randomly selected (again evenly spread across classes) from the MNIST training set. Backward propagation training is run for 250 epochs on the 300 selected images and then tested on the different set of 1000 images. The testing performance of the backward propagation trained ANN becomes the fitness of the CPPN for HyperNEAT. After evolution completes, the generation champions are evaluated on the MNIST testing set. In the HyperNEAT for feature learning case, the backward propagation trained ANN learns from the full MNIST training set.

\begin{figure}[t]
\begin{center}
\centering
\includegraphics[width=0.8\linewidth]{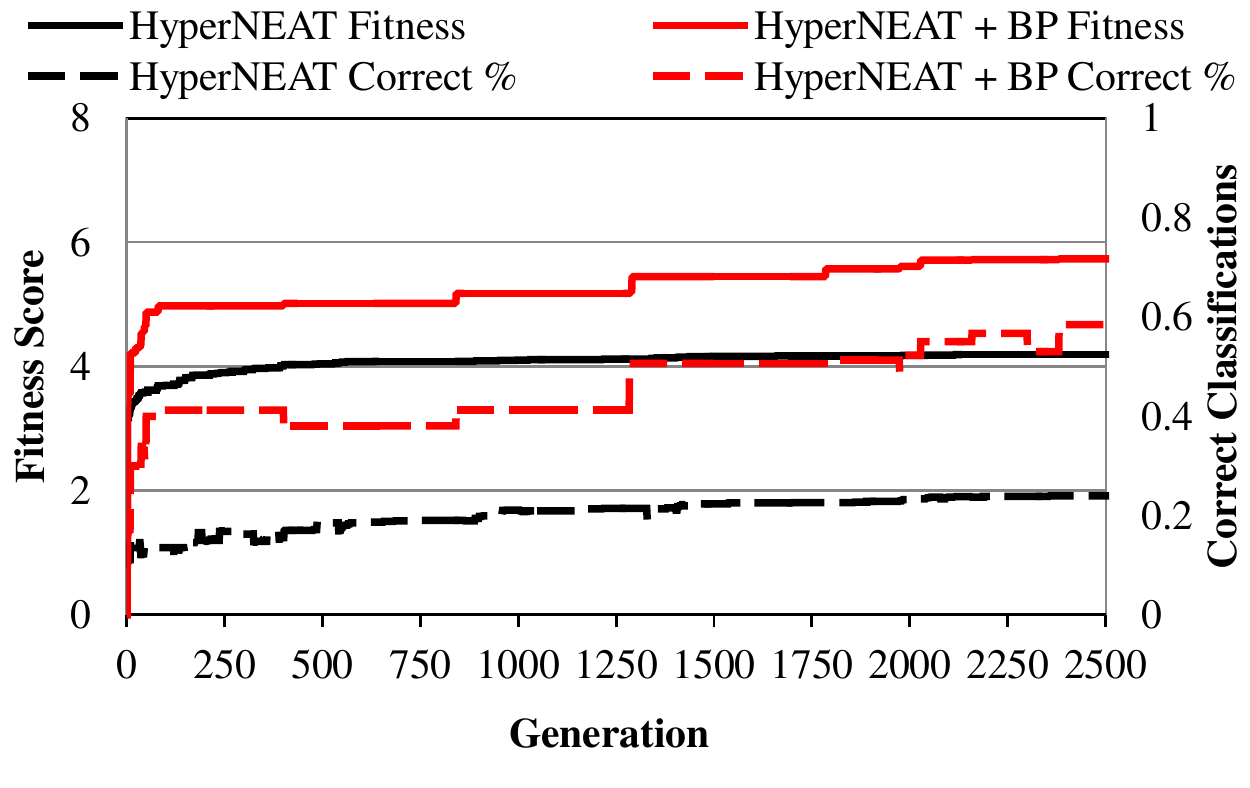}
\end{center}
\vspace{-18pt}
\caption{\textbf{HyperNEAT Performance with Traditional ANN Architecture.} HyperNEAT can learn to classify images by itself; however, learning quickly plateaus at a low performance and then learning slows, reaching a fitness score of $4.2$ and testing performance of $23.9\%$ correct classifications. By acting as a feature learner, HyperNEAT does not plateau and achieves improved performance over HyperNEAT alone, finishing training with a fitness score of $5.7$ and a testing score of $58.4\%$. Thus HyperNEAT as a feature learner is more promising for than HyperNEAT alone. 
\vspace{0.0in}
}
\label{tradann:}
\end{figure}

\indent As seen in figure \ref{tradann:}, HyperNEAT by itself quickly plateaus at a particular performance level and only gradually learns, achieving an average fitness score of $4.2$ by the end of training. The accuracy of these trained solutions on the testing data is also not impressive, achieving only a $23.9\%$ correct classifications. By applying HyperNEAT as a feature learner and allowing backward propagation to train, both fitness score during training and classification correctness during testing increase to $5.7$ and $58.4\%$, respectively. It is interesting to note that in both cases, the fitness score and the correct classifications are not completely correlated, that is, improvements in the fitness score can lead to decreases in classification accuracy.

\begin{figure}[t]
\begin{center}
\centering
\includegraphics[width=0.8\linewidth]{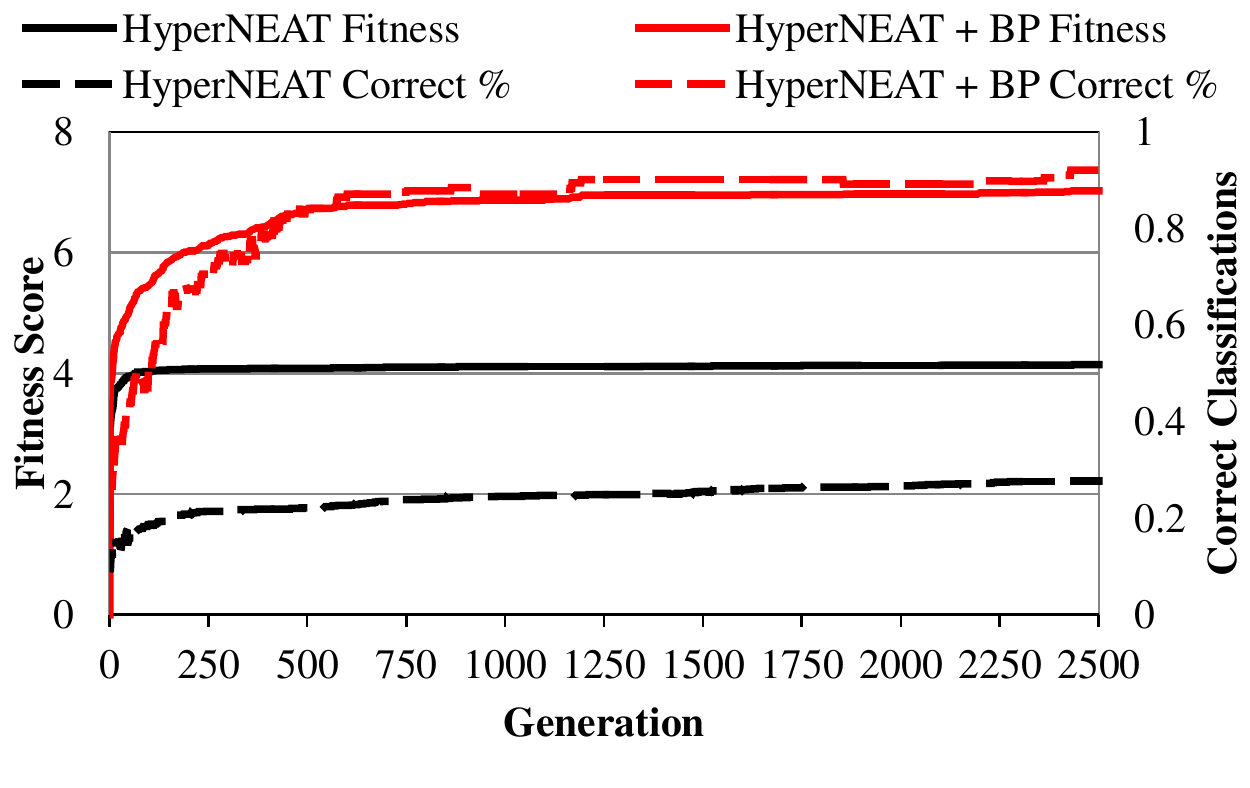}
\end{center}
\vspace{-18pt}
\caption{\textbf{HyperNEAT Performance with CNN Architecture.} The CNN architecture has a small effect on HyperNEAT only performance, lowering the fitness performance plateau to $4.1$, but increasing testing performance of $27.7\%$ correctness. Shifting to CNN architecture significantly improves HyperNEAT's ability as a feature learner, allowing HyperNEAT to find features that achieve a fitness score $7.0$ and a testing score of $92.1\%$. Thus HyperNEAT can learn effective levels of performance given careful selection of the ANN substrate. 
\vspace{0.0in}
}
\label{cnn:}
\end{figure}

\begin{figure}[t]
\begin{center}
\centering
\includegraphics[width=0.7\linewidth]{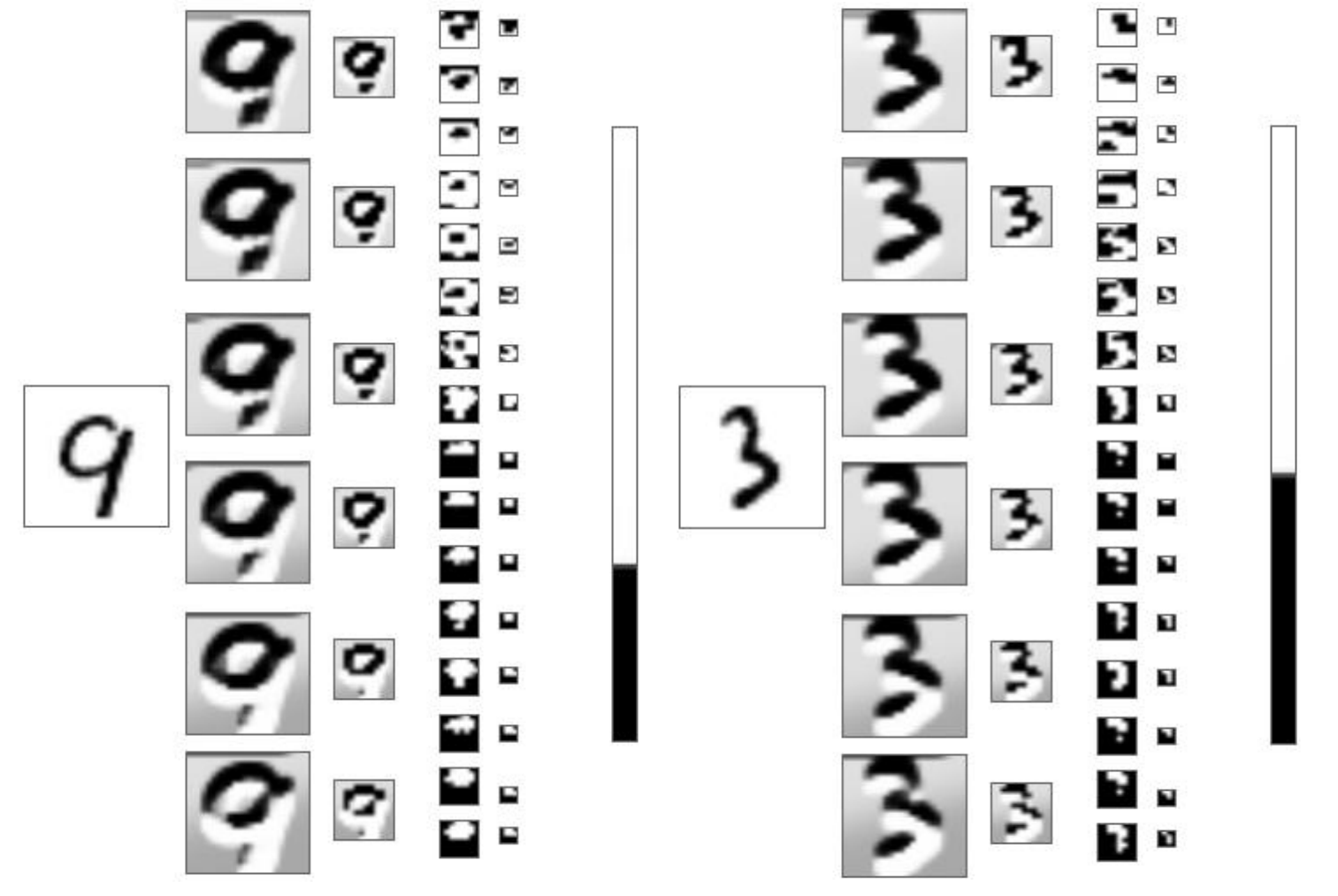}\\
\mbox{(a)\hspace{0.33\linewidth}(b)}
\end{center}
\vspace{-8pt}
\caption{\textbf{Visualization of Feature Maps Generated by HyperNEAT.} Example feature maps are shown for the digits nine (a) and three (b). Interestingly, there is a distinct pattern along the feature dimension (top to bottom), demonstrating patterns in the geometry of the feature maps.
\vspace{0.0in}
}
\label{visual:}
\end{figure}

\indent Changing HyperNEAT to learn the weights of a CNN architecture, rather than the normal ANN architecture of HyperNEAT, does change performance (figure \ref{cnn:}). In this case, HyperNEAT by itself plateaus faster at a lower fitness level, achieving an average fitness score of $4.1$ by the end of training, but the accuracy of these trained solutions on the testing data improves to $27.7\%$ correct classifications. On the other hand, feature learning with HyperNEAT on a CNN architecture significantly improves both metrics, achieving a $7.0$ fitness score and a $92.1\%$ correct classifications. An example of the feature maps generated by these HyperNEAT champions can be seen in figure \ref{visual:}.

\section{Discussion and Future Work}

\indent Evolution is a significant factor in the creation of biological systems, including the visual cortex \cite{le::2013,gliga::2005}. However, biological neural networks are often deep architectures and conventional NeuroEvolution approaches have been challenged in effectively training ANNs order of magnitude smaller than those found in nature. Emerging research from generative and developmental systems has provided an answer in the form of HyperNEAT. HyperNEAT can effectively learn weight patterns for an ANN substrate by training CPPNs, an indirect encoding that computes weights as a function of geometry. The challenge for HyperNEAT is that, by training an indirect encoding, the ability to control precise weights is diminished, thereby creating difficulties in making fine adjustments for tasks such as classification. That is, a change in the indirect encoding will cause changes across the entire weight pattern, when a change to a single weight is needed. This challenge can be addressed by applying HyperNEAT as a feature learner for another ML approach that then makes the fine adjustments for the task, as shown in this paper. Indeed, the results in this paper demonstrate that HyperNEAT successfully learn features to train a backward propagation trained ANN. 

\indent Interestingly, changing to the convolutional neural network architecture had a significant impact on HyperNEAT's performance. 
Two interesting questions arise: (1) Are there more effective architecture choices? and; (2) How do can they be discovered? A path to answering these questions may be an extension of HyperNEAT known as Evolvable Substrate HyperNEAT (ES-HyperNEAT; \cite{Risi2012}). In ES-HyperNEAT, the ANN substrate is not defined a priori (except for inputs and outputs); instead, the pattern generated by the CPPN determines the placement neurons in the hidden layers.

\indent An exciting implication of this work is that evolution provides a means to learn a representation that is well-suited for the target machine learning approach. Through reinforcement learning, evolution can measure fitness as a function of how well an approach performs on the generated representation. 
Because the fitness measure does not depend on a particular error signal or gradient, multiple metrics can be incorporated. For example, the results in the paper incorporated several classification characteristics into the fitness function, such as measures of true positives, true negatives, false positives, and false negatives. By combining measures, evolution can explore multiple paths through the different metrics. In addition, the many candidate solutions that evolution generates will perform differently from each other and these solutions may be combined to create an enhanced feature set. Finally, because the fitness measure can be based upon performance of the trained solution on data not seen by ML approach, the discovered features may encourage learning that generalizes. Thus this approach provides anopportunity to encourage representations that enhance generalizability. 

\section{Conclusion}
\indent This paper investigated deep learning through NeuroEvolution. While NE has been limited in the size of ANNs that could be effectively trained in the past, novel algorithms that operate on indirect encodings, such as HyperNEAT, allow effective evolution of large ANNs. Prior work with HyperNEAT has shown promise in simple vision tasks that operate on a raw visual field, but with non-deep architectures. By itself, HyperNEAT struggles to find ANNs that perform well in image classification; however, HyperNEAT demonstrates an effective ability to act as a feature extractor by being combined with backward propagation. Thus HyperNEAT provides a potentially interesting path for combining reinforcement learning and supervised learning in image classification, as evolution and lifetime learning combine to create the capabilities in biological neural networks. 
\subsubsection*{Acknowledgments}
\indent This work was supported and funded by the SSC Pacific Naval Innovative Science and Engineering (NISE) Program.
\vspace{-8pt}

\bibliographystyle{ieeetr}
\footnotesize
\bibliography{paper,cognet,financial,hertz.refs,multiagent,n,nn,nnstrings,nnvita,oe_refs,temporal,ucf,som_lvq,dissertation}

\end{document}